\documentclass[conference,a4paper]{IEEEtran}

\IEEEoverridecommandlockouts
\usepackage{cite}
\usepackage{amsmath,amssymb,amsfonts}
\usepackage{algorithmic}
\usepackage{graphicx}
\usepackage{textcomp}
\usepackage{xcolor}
\def\BibTeX{{\rm B\kern-.05em{\sc i\kern-.025em b}\kern-.08em
    T\kern-.1667em\lower.7ex\hbox{E}\kern-.125emX}}

\usepackage[table]{xcolor}
\usepackage{booktabs}

\usepackage{amsthm}
\usepackage{multirow}
\usepackage{algorithm}
\usepackage{algorithmic}
\usepackage{placeins}
\usepackage{adjustbox}
\usepackage{array} % for >{\centering\arraybackslash}
\theoremstyle{definition}

\newtheorem{definition}{Definition}

\begin{document}

\title{Dissecting Representation Structure in \\ Vision Transformers: A Rigorous Architectural Study}

\author{\IEEEauthorblockN{Kim-Cuc Nguyen}
\IEEEauthorblockA{\textit{Temasek Laboratories} \\
\textit{Singapore University of Technology and Design}\\
Singapore \\
kimcucnguyen.cs@gmail.com}
\and
\IEEEauthorblockN{Ngai-Man Cheung}
\IEEEauthorblockA{\textit{Temasek Laboratories} \\
\textit{Singapore University of Technology and Design}\\
Singapore \\
ngaiman\_cheung@sutd.edu.sg}
}

\maketitle

\begin{abstract}
Representation structure is crucial for understanding Vision Transformer (ViT) architectures and their generalization behavior. However, prior studies neither isolate nor analyze module-level features nor investigate how their interactions contribute to performance estimation. In this work, we conduct the first rigorous analysis of feature information across diverse architectural scales, empirically uncover the relationship between ViT representation and generalization behavior, and leverage these insights to guide efficient ViT design. Our contributions are fivefold: Across diverse architectural scales, 1) We identify feature collapse at initialization, which leads to redundancy, and propose a reduction scheme to mitigate this issue. 2) We quantify feature information using entropy and the minimum eigenvalue, demonstrating that these metrics serve as reliable indicators for generalization prediction. 3) We show that feature in the token space provides a more faithful representation than those in embedding space. 4) We discover an unexpected finding: features produced by linear submodules within ViT layers are critical for the prediction of generalization performance. 5) Our proposed proxy improves the correlation ranking by 18-48\% over prior baselines and can effectively identify ViT architectures that achieve higher accuracy at lower or comparable computational cost.
\end{abstract}

% \begin{IEEEkeywords}
% component, formatting, style, styling, insert.
% \end{IEEEkeywords}


\section{Introduction}
\label{sec:intro}

The Vision Transformer (ViT) \cite{dosovitskiy2020image} has achieved promising performance in visual tasks \cite{tu2022maxvit, ali2021xcit, dong2022cswin}. The critical components of ViT have been considered to be multi-head self-attention and the multi-layer perceptron (MLP). These specific modules enable the ViT architecture to achieve superior performance, but also result in less interpretable representations. By understanding the specific feature information in ViT, we can design more flexible architectures with stronger generalization ability.

Generalization potential refers to a model’s inherent capacity to generalize prior to training, providing a way to estimate its expected performance without optimization. Although it remains a conundrum, it is a valuable concept for effectively designing ViT architectures.
Moreover, to facilitate the adaptation and application of ViT, Neural Architecture Search (NAS) \cite{chen2021autoformer, su2022vitas, gong2022nasvit, yan2024masked, wang2023prenas, chen2021searching, liu2022uninet, ding2021hr, tang2023elasticvit} has been introduced to automatically discover optimal ViT architectures within a predefined search space, aiming to achieve the best possible accuracy under computational constraints. However, existing ViT NAS methods demand extensive computational resources, as they require training a large Supernet, evaluating numerous sub-architectures, and selecting the optimal one. To mitigate this cost, training-free performance proxies \cite{zhou2022training, wei2024auto, zhou2024training} can be used to estimate the performance of ViT architectures without exhaustively training all candidates in the search space.

\textbf{Research gap.} Existing studies primarily focus on overall architectural design rather than analyzing the specific roles of individual modules. While several proxy-based methods have been proposed to predict network performance, most are adapted from CNN proxies \cite{lee2018snip, wang2023prenas, li2023zico, jiang2024meco, wang2020picking}, which do not generalize well to ViT architectures due to their distinct structural properties. Current ViT proxies \cite{zhou2022training, wei2024auto, zhou2024training, mellor2021neural} typically operate in the weight space or treat the entire architecture as a single entity, overlooking the contribution of individual modules or features. Moreover, the high dimensionality and depth of ViT features make it challenging to quantify their information effectively. 

\begin{figure*}[h]
\begin{adjustbox}{width=0.8\textwidth,center}

\includegraphics[width=0.95\linewidth]{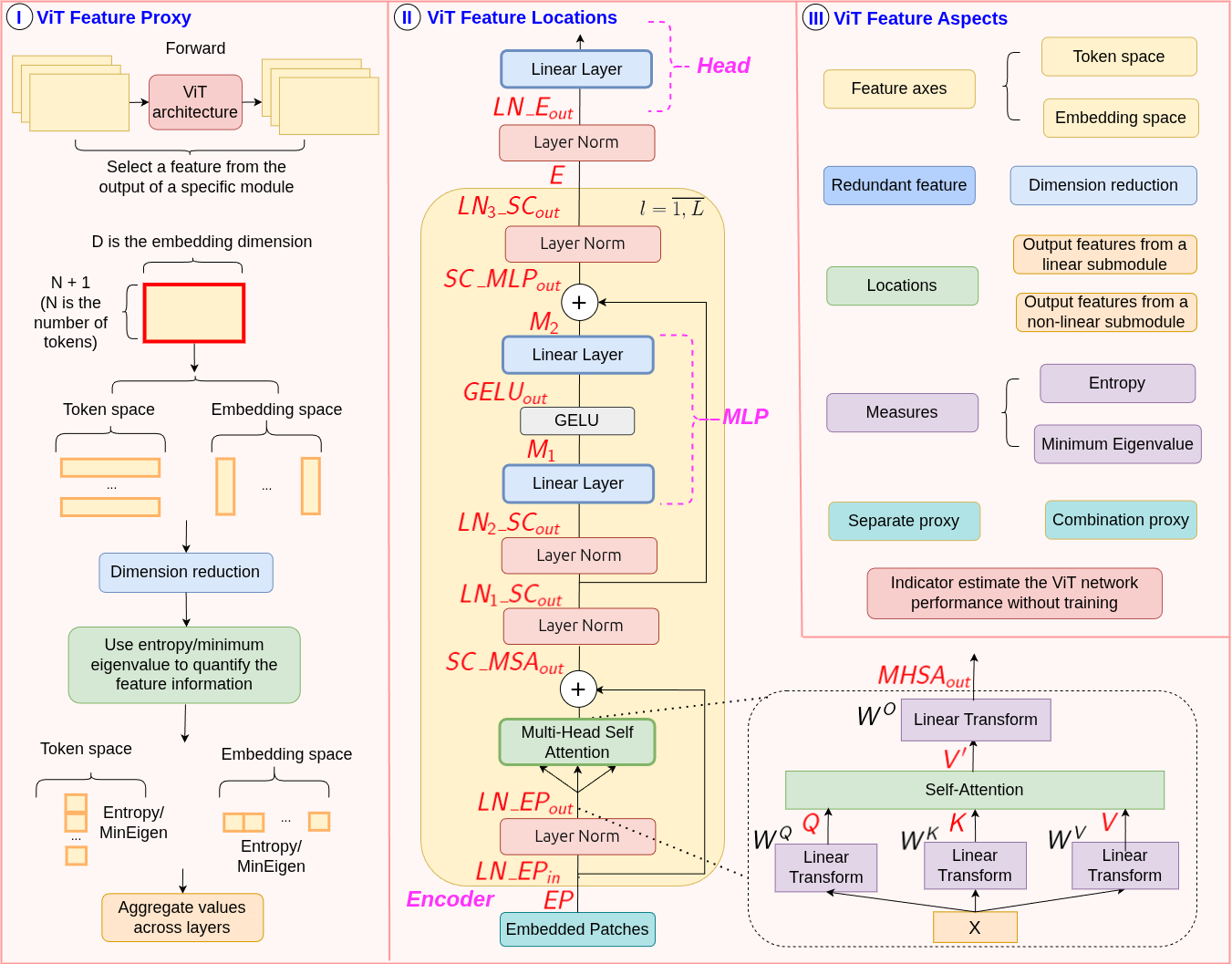}
\end{adjustbox}

\caption{Overview of our proposed framework for rigorous ViT representation analysis.
\textbf{(I)} Procedure for computing feature proxies from specific modules (Sec.~\ref{sec:vit_feature_proxies}).
\textbf{(II)} Visualization of feature locations across ViT modules, with red text indicating feature types and extraction points. 
\textbf{(III)} Analysis of various feature aspects (Sec. \ref{sec:investigation}). We observe feature collapse at initialization (Sec. \ref{sec:feature_collapse}) and propose a reduction scheme to mitigate redundancy. In Tab. \ref{tab:entropy_mineigen_token_embedding}, \textbf{we demonstrate that entropy and the minimum eigenvalue effectively capture feature informativeness, with features output of linear submodule showing strong correlation with test accuracy}, particularly in the token space. The proposed optimal feature proxies predict ViT performance without training, improving correlation ranking and achieving higher accuracy with lower computational cost (Sec. \ref{sec:compare_corr_sota_proxies}, \ref{sec:application_short}.)}

\label{fig:overview}
\vspace{-0.5cm}
\end{figure*}

\textbf{In this paper,} we address existing research gaps by \textit{proposing a framework and conducting the first analytical study of different types of feature information across various aspects of ViT architectural scales}. To mitigate feature collapse, we introduce a reduction scheme that effectively compresses high-dimensional features and streamlines computation. Furthermore, we quantitatively characterize feature information using entropy and the minimum eigenvalue, and show that the outputs of linear submodules are both critical and ubiquitous for generalization prediction.

\section{Related Work}
\label{sec:related_work}

\textbf{Feature collapse.} \cite{laurent2023feature} is among the first works to explore its features in detail, explaining them through the concept of word representations. Importantly, the authors demonstrate that words belonging to the same concept receive identical representations, a phenomenon they define as feature collapse. Using ranking of embedding to represent the information without the join of label \cite{garrido2023rankme} has been used in self-supervised learning. \cite{zhai2023stabilizing} introduce the concept of entropy collapse. In our work, we observe the feature collapse at the initialization of the ViT.

\textbf{Feature selection in ViT.}
Feature selection in ViTs involves identifying the most informative components of the features extracted by the model. Effective feature selection can reduce computational cost, improve generalization, and enhance model interpretability.
\cite{rao2021dynamicvit} proposed a method to prune redundant tokens based on the input by the design prediction module to identify the important token from current features. \cite{yin2022vit} introduced to use an adaptive token mechanism to reduce the number of token processed in ViT. \cite{tang2022patch} proposed removing uninformative patches by identifying the most effective patches in the final layer and using this information for patch selection in earlier layers.

\textbf{Proxy for ViT.} \cite{zhou2022training} proposed measuring the estimated performance of ViT architecture via synaptic diversity and synaptic saliency. \cite{wei2024auto} proposed a method to search for a ViT proxy that is generalizable for various domains and datasets. \cite{wei2023tvt} proposed ViT proxy search for tiny datasets with knowledge distillation. 

\section{Preliminaries}
\label{sec:preliminaries}

\begin{definition}[Shannon Entropy] 
\label{def:entropy}
    The entropy \(H(X)\) of a discrete random variable \(X\) with possible outcomes \(x_1, x_2, \dots, x_n\) and probability function \(P(X)\) is defined as:
    \vspace{-0.2cm}
    \begin{equation}
        \label{eq:entropy}
        H(X) = - \sum_{i=1}^{n} P(x_i) \log P(x_i)
    \end{equation}
    \vspace{-0.2cm}
\end{definition}

\vspace{-0.5cm}

\begin{definition}[Pearson Correlation Matrix]

Let $\mathbf{X} \in \mathbb{R}^{N \times D}$ be a feature matrix, where each column $X_i$ represents one feature vector. The Pearson correlation matrix $\mathbf{P} \in \mathbb{R}^{D \times D}$ is defined as:
\vspace{-0.2cm}
\begin{equation}
\label{eq:pearson_corr}
P_{ij} = \frac{\mathbb{E}[(X_i - \mathbb{E}[X_i])(X_j - \mathbb{E}[X_j])]}{\sigma_{X_i} \, \sigma_{X_j}}
\end{equation}
\vspace{-0.2cm}
where $\sigma_{X_i}$ is the standard deviation of $X_i$.

\end{definition}

\section{Vision Transformer Feature Proxies}
\label{sec:vit_feature_proxies}
In our analysis, we investigate which types of feature and their combinations are the most suitable as indicators for predicting ViT performance. To address this, we first define the core concepts and introduce our proposed methods as follows:

\textbf{Feature Compatibility.} It refers to how features are jointly connected and how their combinations work without conflict. While previous ViT studies assume the whole network, skip connection, or later layers contain more information, table \ref{tab:entropy_mineigen_token_embedding} show that this assumption does not hold for ViT generalization.

\textbf{Feature axes.} Given an input image $X \in \mathbb{R}^{C \times H \times W}$, it is divided into non-overlapping patches, each flattened and linearly projected into a $D$ dimensional embedding. A positional embedding and a learnable class token are then added, forming $X \in \mathbb{R}^{t_{\text{dim}} \times e_{\text{dim}}}$, where $t_{\text{dim}} = N + 1$ is the token dimension and $e_{\text{dim}} = D$ is the embedding dimension. For feature analysis, we treat rows as variables in the token space and columns as variables in the embedding space.

\textbf{ViT Feature.} ViT architecture consists of an Encoder,
Patches, and MLP head, illustrated in Fig.~\ref{fig:overview}. For each module type, we extract one representative feature from its output during a forward pass on a batch of data. Starting from the input, $EP$ denotes the feature after patch embedding. In the Encoder, repeated over $L$ layers, we obtain features from all layers within each module. $LN\_EP_{in}$ and $LN\_EP_{out}$ represent features before and after the first LayerNorm, respectively. The query, key, and value features, denoted as $Q$, $K$, and $V$, are concatenated into $QKV = XW^{QKV}$ for efficiency. $V'$ denotes the feature after self-attention, while $MHSA_{out}$ is the multi-head attention output. $SC\_MSA_{out}$ represents the feature after the skip connection following multi-head self-attention. Next, $LN_1\_SC_{out}$ and $LN_2\_SC_{out}$ represent the features after the first and second LayerNorm operations following the skip connection, respectively. Within the MLP, $M_1$ and $M_2$ denote features from the first and second linear layers, and $GELU_{out}$ is the activation output. $SC\_MLP_{out}$ is the feature after the skip connection following the MLP module, and $LN_3\_SC_{out}$ denotes the feature after the subsequent LayerNorm operation. Finally, $E$ represents the encoder output, and $LN\_E_{out}$ is the feature before the ViT head. Feature combinations are denoted using the plus operator (e.g., $M_1 + M_2$).

\textbf{Feature dimension reduction.}
Representation information measures are computed on $X'$, and the results are scaled by $\tfrac{t_{\text{dim}}}{u}$ in the token space or $\tfrac{e^l_{\text{dim}}}{u}$ in the embedding space to maintain consistency with the original dimensionality. This weighting compensates for dimensionality reduction and ensures that the final information metric reflects the contribution of the entire feature space.

\vspace{2pt}
\begin{algorithm}
\caption{Compute ViT Feature Proxy}
{\small
\begin{algorithmic}[1]
    \label{alg:vit_feature_reduction_proxy}
    \REQUIRE Feature type $f$; feature map $X \in \mathbb{R}^{t_{dim} \times e^l_{dim}}$ corresponding to a single sample in the batch, extracted from module $f$ at layer $l$; reduction dimension $u$; proxy indicator $\mathcal{I} \in \{\text{Entropy}, \text{MinEigen}\}$

    \STATE $proxy \gets 0$
    \FOR{$l \gets 0$ \TO $L$}
        \IF{dim is token}
            \STATE $X=\{x_1, ..., x_{t_{dim}}\}$, $x_i \in \mathbf{R}^{e^l_{dim}}$
            \STATE $\alpha_l \gets \frac{t_{dim}}{{u}}$
        \ELSE
            \STATE $X=\{x_1, ..., x_{e^l_{dim}}\}$, $x_i \in \mathbf{R}^{t_{dim}}$
            \STATE $\alpha_l \gets \frac{e^l_{dim}}{{u}}$
        \ENDIF
        \STATE $X' \gets \text{Reduce}(X, u, \mathcal{I})$ \COMMENT{select $u$ elements using $\mathcal{I}$-specific rule} (see Eqs.~\eqref{eq:reduce_entropy},~\eqref{eq:reduce_corr_s})
        \STATE $p_l \gets \alpha_l \cdot \mathcal{F}_{\mathcal{I}}(X')$ (see Eqs.~\eqref{eq:f_entropy},~\eqref{eq:f_mineigen})
        \STATE $proxy \gets proxy + p_l$

    \ENDFOR
    \RETURN $proxy$

\end{algorithmic}
}
\end{algorithm}
\vspace{2pt}

\textbf{Entropy Proxy. } 
Let $H(x_i)$ denote the entropy of a feature vector $x_i$. For each row or column of the feature matrix $X$, we compute $H(x_i)$ and select the $u$ vectors with the lowest entropy to form a reduced representation $X'$. 
\vspace{-0.2cm}
\begin{equation}
    \label{eq:reduce_entropy}
    X' = \text{Select } u \text{ vectors from X with lowest } H(x_i)
\end{equation}

The entropy proxy function $F_{\mathcal{I}}$ is then computed from the reduced matrix $X'$.
\vspace{-0.2cm}
\begin{equation}
\label{eq:f_entropy}
    F_{\mathcal{I}}(X') = \sum_{i=1}^{u} H(x_i)
\end{equation}

\textbf{Minimum Eigenvalue Proxy.} We compute the Pearson correlation matrix $P(X)$ (Eq. \ref{eq:pearson_corr}) to measure the linear correlation 
between feature vector variables $x_i$ and $x_j$. We define the row-wise correlation score $s_i$, which sum the absolute correlations in each row of $P(X)$.

\vspace{-0.2cm}
\begin{equation}
    s_i = \sum_j \lvert P(X)_{ij} \rvert
\end{equation}
\vspace{-0.5cm}

\vspace{-0.5cm}

\begin{equation}
    \label{eq:reduce_corr_s}
    X' = \text{Select } u \text{ vectors with lowest } s_i
\end{equation}
\vspace{-0.5cm}

\vspace{-0.2cm}
\begin{equation}
\label{eq:f_mineigen}
    F_{\mathcal{I}}(X') = \lambda_{\min}\big(P(X')\big)
\end{equation}
\vspace{-0.5cm}

$\lambda_{\min}(P)$ is proportional to the degree of feature diversity and independence, where higher diversity correlates with better classification accuracy, stronger generalization, and more stable training in ViTs.

\section{Discovering the Impact of Feature Information on ViT Generalization Predictors}
\label{sec:investigation}

\textbf{Procedure.} Our framework and analysis components are illustrated in Fig.~\ref{fig:overview}. We meticulously designed the experimental testbed across different architectural scales. Specifically, we sampled 1,000 ViT architectures within the 5–7 million parameter range from AutoFormer-Tiny \cite{chen2021autoformer}, extracted their architectural information, and stored it as an API. Additionally, we collected the test accuracy of these 1,000 architectures using pretrained weights on the ImageNet-1K dataset \cite{russakovsky2015imagenet}. The test accuracy of subnets inheriting weights from the supernet achieves comparable results to those trained independently \cite{chen2021autoformer, yu2020bignas, wang2021alphanet, wang2021attentivenas}. We propose a unified framework for computing ViT feature proxies, as illustrated in Algorithm~\ref{alg:vit_feature_reduction_proxy}, and introduce two measures: the \textit{Entropy Proxy} and the \textit{Minimum Eigenvalue Proxy}.

\textbf{Metrics.} We computed the Spearman \cite{jh2005spearman} and Kendall~\cite{abdi2007kendall} correlations between proxy values and test accuracy to evaluate the effectiveness of representation information in capturing ViT generalization performance.

\subsection{Investigating Feature Collapse at Initialization and Feature Information for ViT Generalization Prediction}
\label{sec:feature_collapse}
\FloatBarrier

\begin{figure}[h]
\centering

\includegraphics[trim=0 0 0 0, clip, width=0.95\linewidth]{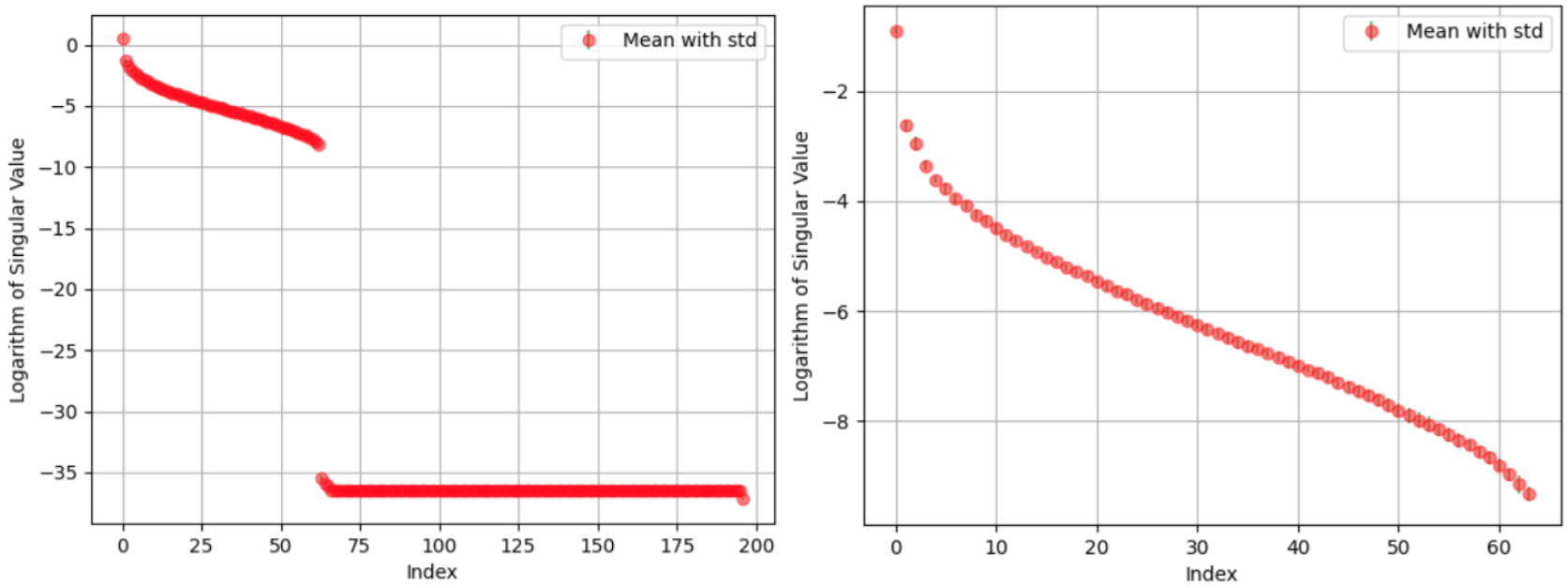}
\caption{Logarithm of the singular value spectrum of the feature $Q$ in the token space (left) and embedding space (right). The x-axis denotes variable indices, and the y-axis shows the mean and standard deviation of the logarithmic singular values across all layers. \textbf{Most singular values are near zero, indicating a low-rank structure and feature collapse.}}
\label{fig:q_token_embedding_features}
\end{figure}

\begin{table}[h]
    \tiny
    \caption{Spearman’s $\rho$ and Kendall’s $\tau$ correlations between test accuracy and each of the \textbf{entropy} and \textbf{minimum eigenvalue} proxies are computed for 1,000 ViT architectures (5–7M parameters) across 26 feature types. These proxies are calculated in both token and embedding spaces for individual and combined features, with dimension reductions $u = 1$. \textbf{Bold} values indicate correlations higher than those of previous proxies in Table~\ref{tab:token_entropy_combination_AFTiny_Small_Base}. \textbf{The entropy and minimum eigenvalue proxies effectively quantify feature information as indicators of ViT generalization. Features output by linear submodules ($Q$, $K$, $V$, $V'$, $MHSA_{out}$, $M_1$, $M_2$) and their combinations show strong correlations with test accuracy, whereas features output by non-linear submodules exhibit significantly weaker correlations. Proxies computed in the token space achieve higher correlations than those in the embedding space.}}

    \centering
    \scriptsize
    
    \begin{adjustbox}{width=0.95\columnwidth,center}

        \begin{tabular}{c|c|c|c|c|c|c|c|c|c}
        \multicolumn{2}{c|}{\multirow{3}{*}{Proxy}} 
        & \multicolumn{4}{c|}{Entropy} 
        & \multicolumn{4}{c}{Minimum Eigenvalue} \\ 
        \cline{3-10}

        \multicolumn{2}{c|}{} 
        & \multicolumn{2}{c|}{Token} 
        & \multicolumn{2}{c|}{Embedding} 
        & \multicolumn{2}{c|}{Token} 
        & \multicolumn{2}{c}{Embedding} \\ 
        \cline{3-10}

        \multicolumn{2}{c|}{} & $\rho$ & $\tau$ & $\rho$ & $\tau$ & $\rho$ & $\tau$ & $\rho$ & $\tau$ \\ \hline

        \multirow{12}{*}{Encoder} & $Q$  & \textbf{0.745}  & \textbf{0.509} & \textbf{0.732} & \textbf{0.480} &
        \textbf{0.796} & \textbf{0.636}  &  \textbf{0.794}  & \textbf{0.633} \\

        & $K$ & \textbf{0.733} & \textbf{0.485} 
        & \textbf{0.738} & \textbf{0.495} 
        & \textbf{0.792} & \textbf{0.630} 
        & \textbf{0.797} & \textbf{0.637} \\

        & $V$ & \textbf{0.744} & \textbf{0.503}  
        & \textbf{0.718} & \textbf{0.468} 
        & \textbf{0.797} & \textbf{0.637}  
        & \textbf{0.796} & \textbf{0.637}  \\

        & $V'$ & \textbf{0.787} & \textbf{0.578}
        & 0.666 & 0.453  
        & \textbf{0.792} & \textbf{0.628}  
        & \textbf{0.699} & \textbf{0.540} \\

        & $MHSA_{out}$ & \textbf{0.790} & \textbf{0.585}  
        & 0.657 & 0.439 
        & \textbf{0.794} & \textbf{0.633} 
        & \textbf{0.701} & \textbf{0.541}  \\

        & $M_1$ & \textbf{0.844}  & \textbf{0.660}  &  \textbf{0.762} & \textbf{0.571}  &  \textbf{0.795} & \textbf{0.633}  &  \textbf{0.761} & \textbf{0.585}  \\
        
        & $M_2$&\textbf{0.809}  & \textbf{0.599}  & \textbf{0.676} & \textbf{0.463} &  \textbf{0.761} & \textbf{0.589}  &  \textbf{0.698} &  \textbf{0.537} \\
        
        & $GELU_{out}$ & 0.137 & 0.104  & 0.131 & 0.100  &  0.016  & 0.013 &  0.129  & 0.104 \\

        & $SC\_MSA_{out}$ & -0.039  & -0.023  &  -0.056 & -0.038  &  0.008  & 0.007 &  -0.138  & -0.109   \\
        
        & $SC\_MLP_{out}$ & -0.023 & -0.014 &  -0.043 & -0.029 &  0.042 & 0.034  &  -0.138 & -0.109 \\

        & $LN\_EP_{in}$ & -0.024  & -0.014  & -0.050 & -0.033 & 0.021 & 0.017 &  -0.138 &  -0.109  \\
        
        & $LN\_EP_{out}$ & -0.031  & -0.019 & -0.071 & -0.047  &   0.052  &  0.042  &   -0.142  & -0.113 \\       

        & $LN_1\_SC_{out}$ & -0.039  & -0.023 & -0.056 & -0.038 &  0.008  & 0.007     &  -0.138  & -0.109   \\
        
        & $LN_2\_SC_{out}$& -0.029  & -0.017   & -0.055  &  -0.038 &   0.037 & 0.029     &  -0.139 &  -0.110   \\

        & $LN_3\_SC_{out}$& -0.023 & -0.014   &  -0.043  &  -0.028 &  0.042 & 0.034   & -0.138 &  -0.109  \\

        Patches & EP & -0.192 & -0.158 &  -0.192 & -0.157 &  NA & NA  &  -0.192 &  -0.157   \\
        
        \multirow{2}{*}{MLP Head} & E & -0.055 & -0.034  & -0.088 & -0.059   & 0.076  & 0.061   & -0.126 & -0.096 \\
        
        & $LN\_E_{out}$ & -0.011  & -0.006 &  0.216  & 0.133 &  0.091 & 0.074   &   -0.086 & -0.067  \\
        
        \multicolumn{2}{c|}{$QKV$} & \textbf{0.773}  & \textbf{0.544}  & \textbf{0.720} & \textbf{0.496} &  \textbf{0.795}  & \textbf{0.634} & \textbf{0.734} & \textbf{0.529} \\
        
        \multicolumn{2}{c|}{$M_1 + M_2$} & \textbf{0.849} & \textbf{0.665}   &  \textbf{0.752} & \textbf{0.562} &   \textbf{0.797} & \textbf{0.639}  &  \textbf{0.752} & \textbf{0.576} \\
        
        \multicolumn{2}{c|}{$QKV$ + $M_1 + M_2$}&   \textbf{0.838}  & \textbf{0.652}  &  \textbf{0.837}  & \textbf{0.649} &   \textbf{0.798}  & \textbf{0.640}  & \textbf{0.836} & \textbf{0.654} \\
        
        \multicolumn{2}{c|}{$QKV$ + $MHSA_{out}$} &  \textbf{0.805}  & \textbf{0.604} &  \textbf{0.773}  & \textbf{0.553} &   \textbf{0.798} & \textbf{0.639} & \textbf{0.771} & \textbf{0.567} \\
        
        \multicolumn{2}{c|}{$MHSA_{out} $ + $M_1$ + $M_2$} & \textbf{0.851} & \textbf{0.669} &  \textbf{0.744} & \textbf{0.556}  & \textbf{0.798} & \textbf{0.640}    &  \textbf{0.745} & \textbf{0.570}\\
        
        \multicolumn{2}{c|}{$QKV$ + $MHSA_{out} $ + $M_1$ + $M_2$} &  \textbf{0.856}  & \textbf{0.676} & \textbf{0.759}  & \textbf{0.574}  &  \textbf{0.798} & \textbf{0.640}  & \textbf{0.831} & \textbf{0.650}  \\   

        \multicolumn{2}{c|}{$Q + K + V$} &  \textbf{0.740} & \textbf{0.499}   &  \textbf{0.728} & \textbf{0.477}    &  \textbf{0.798} & \textbf{0.640}  &  \textbf{0.798} & \textbf{0.639}   \\
        
        \multicolumn{2}{c|}{$SC\_MSA_{out} + SC\_MLP_{out}$} & -0.040  & -0.024   & -0.056  & -0.037  & 0.027   & 0.022  & -0.143 & -0.114
        \\
        \hline
    \end{tabular}
\end{adjustbox}
\label{tab:entropy_mineigen_token_embedding}
\end{table}

\vspace{2pt}

\subsection{Comparison with State-of-the-Art Methods}
\label{sec:compare_corr_sota_proxies}

We compare the top three Token-Entropy proxies with reduced dimension $u = 1$: $M_1 + M_2$, $MHSA_{out} + M_1 + M_2$, and $QKV + MHSA_{out} + M_1 + M_2$, against the baselines across 3,000 ViT architectures: Tiny (5-7M), Small (15-19M), and Base (45-47M), with 1,000 architectures in each parameter range. As shown in Table~\ref{tab:token_entropy_combination_AFTiny_Small_Base}, our proxies improve correlation rankings by 18\%, 25\%, and 48\% for the 5–7M, 15–19M, and 45–47M ranges, respectively.

\begin{table}[htbp]  
    \vspace{5pt}
    \caption{Spearman’s $\rho$ and Kendall’s $\tau$ correlation rankings for three sets of 1,000 ViT architectures with parameter ranges of 5–7M, 15–19M, and 45–47M, comparing the top-3 Token-Entropy proxies ($u = 1$) with baselines. \textbf{Our proposed proxies significantly improve the correlation rankings, demonstrating that they are strong indicators of ViT performance.}}
    \centering
    \small
    
    \begin{adjustbox}{width=1\columnwidth,center}

    \begin{tabular}{c|c|c|c|c|c|c}
        \multirow{2}{*}{Proxy} & \multicolumn{2}{c|}{5-7M} & \multicolumn{2}{c|}{15-19M} & \multicolumn{2}{c}{45-47M} \\ 
        \cline{2-7}
         & $\rho$ & $\tau$ & $\rho$ & $\tau$ & $\rho$ & $\tau$ \\ 
        \hline
        SNIP  \cite{lee2018snip} & 0.313 & 0.207 & 0.280 & 0.190 & 0.056 & 0.037 \\
        GraSP \cite{wang2020picking}& -0.101 & -0.066 &  -0.064& -0.043 & 0.023 & 0.015 
        \\
        TE-score \cite{chen2021neural} & -0.319 & -0.219 & -0.084 & -0.057 & -0.106 & -0.072  \\
        NASWOT \cite{mellor2021neural} & 0.382 & 0.278 & 0.232 & 0.162 & 0.243 & 0.171 \\
        DSS \cite{zhou2022training} & 0.622 & 0.439 & 0.468 & 0.315 & -0.119 & -0.079\\
        AutoProxA \cite{wei2024auto} & 0.675 & 0.477 & 0.446 & 0.299 & -0.126 & -0.084 \\
        DSS++ \cite{zhou2024training}  & 0.638 & 0.448 & 0.450 & 0.302 & -0.140 & -0.094\\
        \rowcolor{gray!15} $M_1 + M_2$ & \textbf{0.849} & \textbf{0.665} &\textbf{0.718} & \textbf{0.529} & \textbf{0.368} & \textbf{0.251} \\
        \rowcolor{gray!15} $MHSA_{out} $ + $M_1$ + $M_2$ &  \textbf{0.851} & \textbf{0.669} & \textbf{0.717} & \textbf{0.529} & \textbf{0.360} & \textbf{0.246} \\
        \rowcolor{gray!15} $QKV$ + $MHSA_{out} $ + $M_1$ + $M_2$ & \textbf{0.856}  & \textbf{0.676} & \textbf{0.720} & \textbf{0.533} & \textbf{0.350}  & \textbf{0.239} \\
    \end{tabular}
    \end{adjustbox}
    \label{tab:token_entropy_combination_AFTiny_Small_Base}

\vspace{-2pt}
\end{table}

\section{Efficient ViT Architecture Design}
\label{sec:application_short}
We use the previous proxies and our Token-Entropy proxies with reduced dimension $u=1$ to search for the optimal subnet on the same set of 1,000 ViT architectures within 5-7 parameter ranges. As shown in the table \ref{tab:zero_proxies_acc_AF_5_7M_15_19M}, our proposed proxies successfully identify the optimal subnet, improving accuracy by 0.11\%-0.17\% compared to previous proxies across the 1,000 ViT architectures in the 5–7M ranges, respectively. Moreover, our proxies achieve an 87.5\%–98.6\% reduction in computation time when searching over 1,000 ViT architectures, requiring only 0.06 hours compared to 0.48 – 4.3 hours for previous proxies in the 5–7M range.

\FloatBarrier
\begin{table}[h]

    \centering

    \caption{Results on ImageNet using different proxies to search for the optimal architecture among 1,000 ViT architectures, with the 5–7M parameter ranges}
   
    \begin{adjustbox}{width=0.95\columnwidth,center}
    \begin{tabular}{l|cc|cc}
         Proxy & Param(M) $\downarrow$ & FLOPs(B) $\downarrow$ & Top-1 $\uparrow$ & Time(h) $\downarrow$  \\ \midrule

         SNIP  \cite{lee2018snip}  & 6.9 &  1.7 & 74.89 &  1.00  \\

         GraSP \cite{wang2020picking} & 5.7  & 1.4 & 74.47 &  1.38  \\

         TE-score \cite{chen2021neural} & 5.9 & 1.5 & 74.60 & 4.30 \\

         NASWOT \cite{mellor2021neural} & 6.9 & 1.7 & 74.94 & 0.82 \\

         DSS \cite{zhou2022training} & 6.9 & 1.6  & 75.29 &  0.60 \\

         AutoProxA \cite{wei2024auto} & 6.9 & 1.5  & 75.32 &  0.48 \\

         DSS++ \cite{zhou2024training} & 6.9 & 1.6 & 75.30 & 0.80 \\ 

         \rowcolor{gray!15} $M_1 + M_2$ & \textbf{6.9} & \textbf{1.5} & \textbf{75.41} & \textbf{0.05} \\
         \rowcolor{gray!15} $MHSA_{out} $ + $M_1$ + $M_2$ & \textbf{6.9} & \textbf{1.5}  & \textbf{75.43} & \textbf{0.06}  \\
         \rowcolor{gray!15} $QKV$ + $MHSA_{out} $ + $M_1$ + $M_2$ & \textbf{6.9} & \textbf{1.5} & \textbf{75.38} & \textbf{0.08} \\ \midrule

    \end{tabular}

    \end{adjustbox}
    \label{tab:zero_proxies_acc_AF_5_7M_15_19M}
\end{table}

\section{Conclusion}
\label{sec:conclusion}

In this work, we provide a rigorous analysis of how feature representations relate to ViT generalization, considering feature information, axes, dimensions, and locations. Our proposed feature proxies improve correlation ranking by 18–48\% over baselines across parameter ranges. For efficient ViT architecture design, our method achieves higher accuracy across architectures while significantly reducing training, search, and proxy computation costs. Extensive experiments across different ViT architectural scale confirm the effectiveness and broad applicability of our approach, establishing a principled framework for understanding ViT features, predicting generalization, and enabling efficient ViT design.
Future work applies the findings to design compact models for different tasks
\cite{song2016egocentric,min2021integrating,chandrasegaran2022discovering}.

{
\small
\section*{Acknowledgment}
This research is supported by Temasek Laboratories, Singapore University of Technology
and Design; the National Research Foundation, Singapore under its AI Singapore Programmes (AISG Award No.: AISG2-TC-2022-007); the Agency for Science, Technology
and Research (A*STAR) under its MTC Programmatic Funds (Grant No. M23L7b0021);
the National Research Foundation, Singapore and Infocomm Media Development Authority
under its Trust Tech Funding Initiative. Any opinions, findings and conclusions or recommendations expressed in this material are those of the author(s) and do not reflect the
views of National Research Foundation, Singapore and Infocomm Media Development Authority.
}

{
    \footnotesize
    \bibliographystyle{IEEEtran}
    \bibliography{main}
}

% \bibliography{main}

\end{document}